\documentclass{article}
\usepackage{arxiv}
\usepackage{times}
\usepackage{epsfig}
\usepackage{graphicx}
\usepackage{amsmath}
\usepackage{amssymb}
\usepackage{booktabs}
\usepackage{comment}
\usepackage{afterpage}
\usepackage{pdflscape}
\usepackage{rotating}
\usepackage{multirow}
\usepackage{multicol}
\usepackage{tabularx}
\usepackage{tikz}
\usepackage{enumitem}
\usepackage{subcaption}
\usepackage{pifont}
\usepackage{adjustbox}
\usepackage{hyperref}
\usepackage[flushleft]{threeparttable} 
\usepackage{accents}
\usepackage{amsfonts}
\DeclareRobustCommand{\eg} {\textit{e}.\textit{g}.}
\DeclareRobustCommand{\ie}{\textit{i}.\textit{e}.~}
\DeclareRobustCommand{\etal}{\textit{et~al.}~}

\newcommand{\br}{\mathbf{r}}

\newcommand{\ba}{\mathbf{a}}
\newcommand{\bx}{\mathbf{x}}
\newcommand{\bs}{\mathbf{s}}
\newcommand{\bz}{\mathbf{z}}
\newcommand{\by}{\mathbf{y}}

\newcommand{\bn}{\mathbf{n}}

\newcommand{\bS}{\mathbf{S}}

\newcommand{\EL}{\mathcal{L}}

\newcommand{\ED}{\mathcal{D}}

\newcommand{\EC}{\mathcal{C}}

\newcommand{\Geom}{\mathcal{G}_{Geom}}
\newcommand{\Render}{\mathcal{G}_{R}}
\newcommand{\CT}{\mathcal{G}_{CT}}
\newcommand{\CM}{\mathcal{M}_{C}}

\newlength{\tempdima}
\newcommand{\rowname}[1]
{\rotatebox{90}{\makebox[\tempdima][c]{\textbf{#1}}}}

\AtBeginDocument{%
  }

\title{MUNCH: Modelling Unique 'N Controllable Heads}
\author{Debayan Deb \\
LENS, Inc.\\
  4288 Indian Glen Drive\\
  Okemos, Michigan, USA \\
\texttt{debayan@lenscorp.ai}
  \And
Suvidha Tripathi \\
LENS, Inc.\\
  4288 Indian Glen Drive\\
  Okemos, Michigan, USA \\
  \texttt{suvidha.tripathi@lenscorp.ai}
  \And
Pranit Puri \\
LENS, Inc.\\
4288 Indian Glen Drive\\
Okemos, Michigan, USA \\
\texttt{pranit.puri@lenscorp.ai}
}

\begin{document}
\maketitle

\begin{figure*}[h]
\begin{center}
    \centering
    \includegraphics[width=\linewidth]{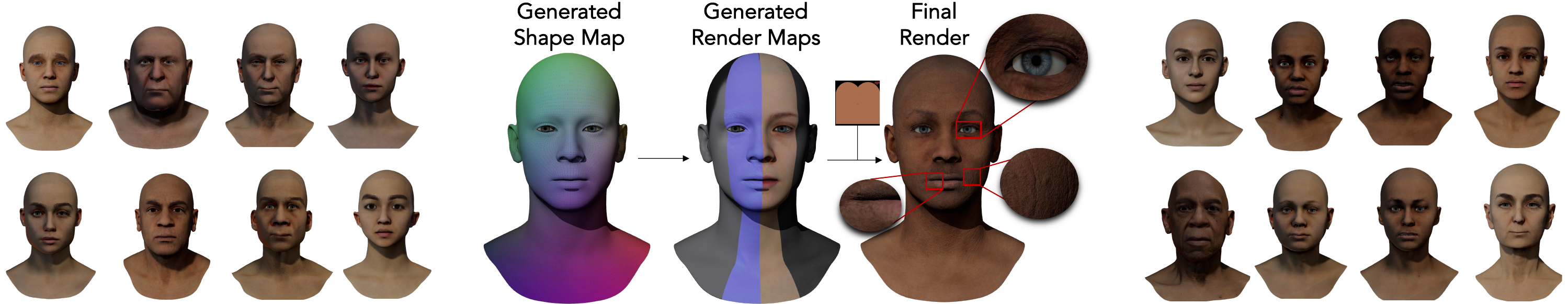}
    \captionof{figure}{Achieving realism in 3D modelling is not a one step process. Currently, 3D game artists in the domain undergo multiple stages, starting from shape sculpting to texturing and then rendering a single character in several months. This work attempts to reduce the manual efforts through AI assisted modeling incorporating user control like age, gender, and race, along with skin color of the 3D human head. Shown here are few examples rendered via our proposed method  along with intermediate outputs from the proposed pipeline (in the middle).}
  \label{fig:teaser}
\end{center}
\end{figure*}
\begin{abstract}
The automated generation of 3D human heads has been an intriguing and challenging task for computer vision researchers. Prevailing methods synthesize realistic avatars but with limited control over the diversity and quality of rendered outputs and suffer from limited correlation between shape and texture of the character. We propose a method that offers quality, diversity, control, and realism along with explainable network design, all desirable features to game-design artists in the domain. First, our proposed Geometry Generator identifies disentangled latent directions and generate novel and diverse samples. A Render Map Generator then learns to synthesize multiply high-fidelty physically-based render maps including Albedo, Glossiness, Specular, and Normals. For artists preferring fine-grained control over the output, we introduce a novel Color Transformer Model that allows semantic color control over generated maps. We also introduce quantifiable metrics called Uniqueness and Novelty and a combined metric to test the overall performance of our model. Demo for both shapes and textures can be found: \url{https://munch-seven.vercel.app/}. We will release our model along with the synthetic dataset. 
\end{abstract}

\keywords{3D reconstruction, Image-based modeling, Mesh processing, Shape analysis, Photogrammetry}

\section{Introduction}

\label{sec:intro}
\begin{figure*}[!t]
    \centering
    \includegraphics[width=0.9\linewidth]{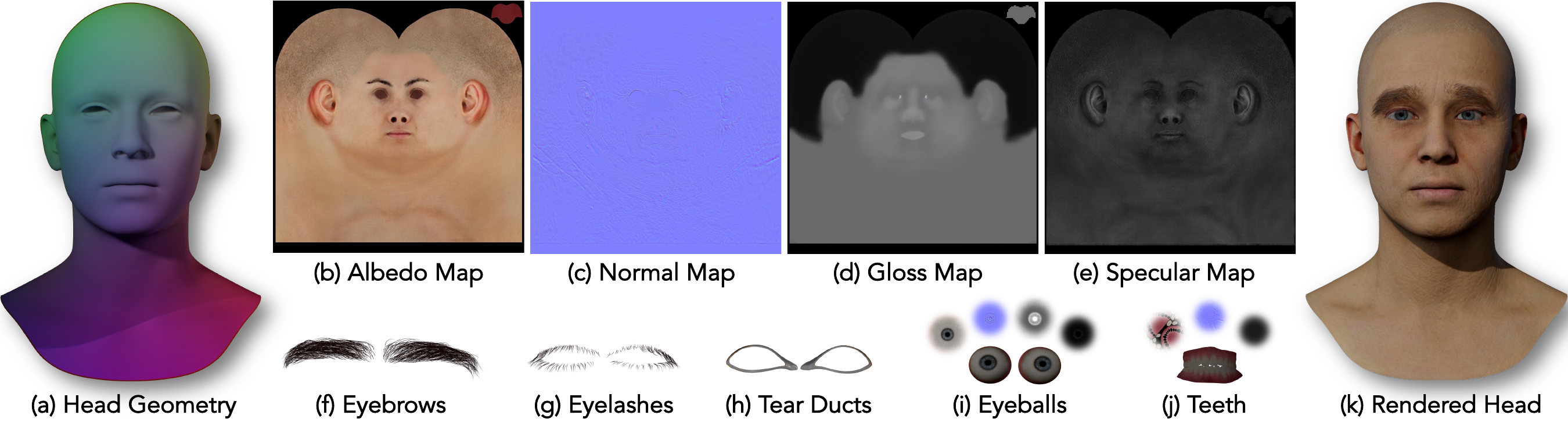}
    \caption{Our generic head model comprises of multiple geometries that all contribute to the final perceptual realism of rendered heads. In addition to (a) the face and neck, our model also incorporates (b) albedo, (c) normal, (d) gloss, and (e), spec. maps along with (f) eyebrows, (g) eyelashes, (h) tear ducts, (i) eyeballs, and (j) teeth. We show the (k) final rendered head with all components put together.}
    \label{fig:components}
\end{figure*}
We all enjoy playing games and watching movies with high-quality 3D effects involving technologies like VFX and CGI. The realism in the characters make us wonder how such character in the game/movie is framed to make it look so real. Other than generated assets, it is very intriguing to naive people that how game avatars and characters can take the frame of a real person who is playing it. What would be more interesting  if we could render our own doppelganger which looks just like us and attend online meetings. What we don't realise is that it takes hours to synthesize or model such 3D characters by graphic designers. With increasing demand for realism in the industry, the job is becoming even harder.

To make the life of the game artist easier, the researchers in the domain of 3D graphics and computer vision focused their thoughts on generating high resolution geometry and visually realistic textures. Recent works like \cite{gecer2019ganfit, lattas2021avatarme++, lattas2020avatarme, li2020learning, wood2021fake, gecer2020synthesizing, gecer2021fast} worked towards making their outputs as close to real person as possible. Their work is remarkable and opens opportunities to render hyper-realistic face models. However, they all lacked in meeting artistic use-cases for building robust 3D characters. 

Artists want maximum control over generated geometry and corresponding texture of human heads. They may require to edit both head geometry and textures after their generation from the automated linear \cite{paysan20093d,li2017learning} or non-linear methods \cite{ranjan2018generating, bouritsas2019neural, taherkhani2022controllable, foti20223d}. While previous literature \cite{murphy2021artist, murphy2020appearance, li2020learning} took attempt to semantically control physical and demographic attributes, their methods lack in presenting a consolidated network design. One architecture should have both disentangled and entangled features as required by game artists. For example, demographic attributes like age, race, and gender when provided as input should output entangled geometry and texture since a person from particular ethnicity have distinct physical face shape and texture color. Whereas, within texture color, their could be a range of color interpolation, like for a mixed race person, the texture color can vary between light to dark skin tone, with corresponding change in color in eyebrows and lips. Thus, an option to control color of the output texture map through the \emph{Color Transformer Model} provides flexibility to game artists to interpolate facial features within a particular demographic cohort. Due to two-stage process, the user, if desires to modify the color gradient of the sample output, could control the face, eyebrow, and lip colour of the 3D model.

Besides realism and control over generated meshes, it is imperative that the generated geometry is highly diverse among all generated samples with in a particular demographic cohort so that the game artist has the flexibility to choose from several options according to the application. Novelty is also an important measure to quantify whether the generated samples are different from the available training set. Otherwise, with low novelty, artist can just choose from the real data.  
\paragraph{Contribution} Prior works in the 3D domain focused on generating high quality renders aiming at realism. In our work, in addition to quality, we have also focused on controllability over 3D assets. We have also addressed the lack of test metrics which can quantify the diversity and uniqueness of the generated 3D assets. These quantifiable metrics gives empirical proof of generalizability of one's model. While the effectiveness of linear models like PCA has already been explored in pioneer works like FLAME and 3DMM, there is a lack in the quantifiable metrics to evaluate the diversity and specificity of their generated heads. To this end, we introduced metrics like 'Uniqueness', and 'Novelty' to evaluate and compare our model with existing state of the art models. 

To the best of our knowledge, we are first to attempt building a hybrid and flexible model favourable for game artists to create their 3D assets (see Fig.~\ref{fig:components}) with maximum diversity, novelty, correlation, realism (high resolution), and control. Please refer to the supplementary file for qualitative evaluation of mentioned characteristics. 

\section{Related work}
\label{sec:rev}

\paragraph{3D shape generation - parametric models} The pioneer work of Blanz and Vetter in 1999 \cite{blanz1999morphable} popularized the use of 3D Morphable Models (3DMM) for generating new meshes or reconstructing 3D faces from single 2D images. 3DMM is a parametric model developed by fitting scans to a multivariate normal distribution based on mean and variance of shape and texture of 200 scans. Many morphable models that cover facial regions like BFM 2009 \cite{paysan20093d}, LSFM \cite{booth20163d}, LYHM \cite{dai20173d}, and BFM2017 \cite{gerig2018morphable} have been proposed since then to generate new identities in geometry. Few of them offer both shape and texture models  \cite{paysan20093d,booth20163d,li2017learning}. These 3DMM models form the basis for applications such as 3D face reconstruction from single images \cite{marriott20213d, yamaguchi2018high, gecer2021ostec, lin2022realistic}. Later full head models like FLAME \cite{li2017learning}, LYHM \cite{dai2020statistical}, UHM \cite{ploumpis2019combining, ploumpis2020towards} were introduced for game artist to bring more flexibility towards choice of building realistic characters, which was not possible with only face models. With full head, the artists could add hairs and head accessories and good texture model could bring more realism to the generated game characters or virtual avatars.
All these methods majorly used either PCA based morphable models or Linear Blend Skinning (LBS) methods for further adding blendshapes to these models.  Although highly feasible for generating new identities, these models pose several limitations as mentioned in Section \ref{limit}
\paragraph{Generative 3D networks: Non-parametric models} 3DMM based facial geometry reconstruction method by \cite{sela2017unrestricted} uses image-to-image translation network \cite{isola2017image} to generate depth maps and correspondence maps. Non-parametric methods involving convolutional operations use Generative Adversarial Networks (GANs) \cite{moschoglou20203dfacegan, gecer2020synthesizing, murphy2021artist, wood2021fake, li2020learning} with 2D shape maps as inputs, and Variational Autoencoders (VAEs) \cite{ranjan2018generating, bouritsas2019neural, gong2019spiralnet++, zhou2019dense, taherkhani2022controllable, foti20223d} use direct 3D meshes as inputs for generating or reconstructing geometry.

GANs offer a non-parametric method to obtain 3D faces by mapping to non-linear space and hence are perceived to be able to model non-linear variations in geometry and textures. Variational Auto Encoders (VAEs) are extensively used for both latent interpolation and reconstruction of geometry and textures. \cite{aliari2023face} published an impressive work using VAEs to allow interactive and fine grained 3D face editing. Methods such as \cite{gecer2020synthesizing, li2020learning} jointly model geometry and textures using known GAN architectures like StyleGAN \cite{karras2020training} and progressive GAN \cite{karras2017progressive}. Slossberg et al. \cite{slossberg2018high} focused more on generating high quality textures than high resolution geometry. They argued through their analysis that the detail geometry has a small impact while a high resolution texture makes a larger impact for rendering realistic characters.
\paragraph{Game quality realism and control} The most realistic looking models till date are achieved by 3D graphics designers and game artists who work hours to generate high quality textures with pore level details. The amount of man hours, efforts, and skill required to bring realism in avatars or game characters speaks for itself the need for developing automatic methods. Murphy~\etal~\cite{murphy2020appearance} proposed generative model that takes specific description of a character and outputs best fitting textures and head shapes. The method allows control over demographic attributes.  \cite{lin2022realistic} creates realistic 3D game avatars from 2D input just like MeInGame \cite{lin2021meingame}, and AvatarMe \cite{lattas2020avatarme}. The authors used 3DMM to create face mesh and then used RBF to transfer the shape of 3DMM face to the game template head mesh. To improve the realism factor, they trained an encoder-decoder network both with Albedo and normal maps.Other methods that achieved game quality like textures are \cite{gecer2019ganfit, gecer2021fast, marriott20213d, murphy2020appearance, murphy2021artist, gecer2020synthesizing, wood2021fake, li2020learning, lin2022realistic,saito2017photorealistic, lattas2020avatarme, lattas2021avatarme++, yamaguchi2018high}. 

\section{Limitations of previous works}
\label{limit}
\begin{figure*}[!t]
    \centering
    \includegraphics[width=0.9\linewidth]{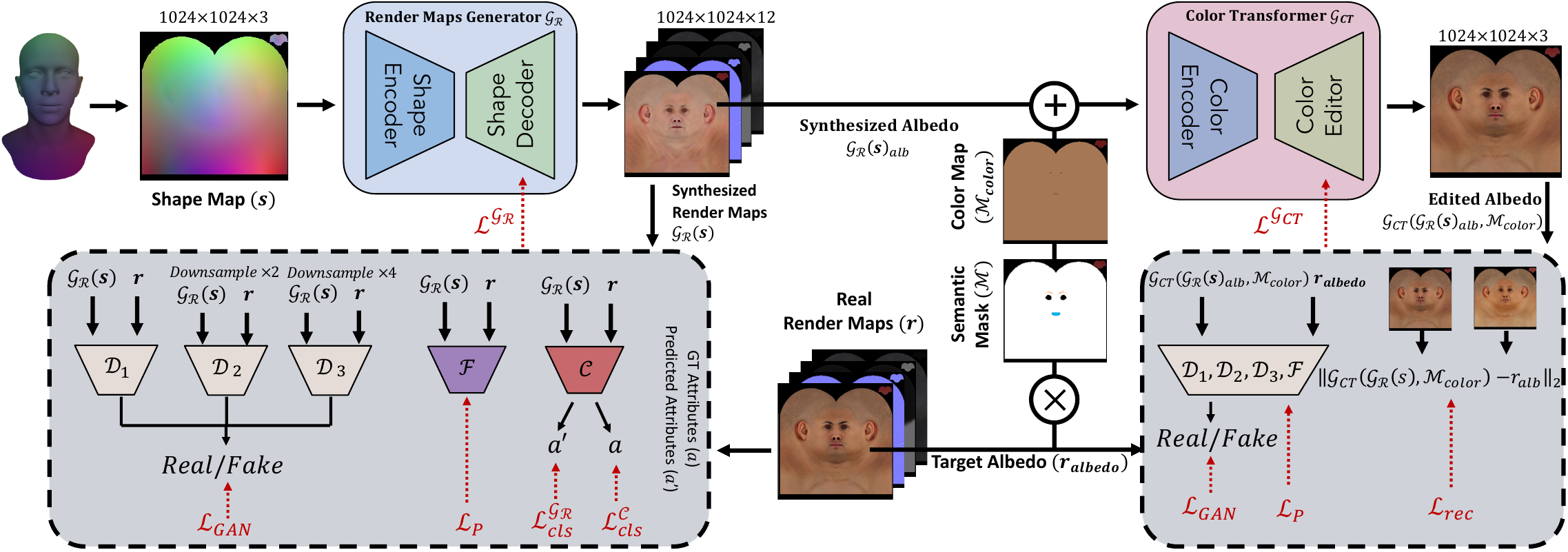}
    \caption{Overview of our proposed training framework. We first cylindrically unwrap shape information and perform barycentric interpolation to obtain dense shape maps. Then, our proposed Render Maps Generator $\Render$ takes shape maps as inputs and outputs as $12$-channel render map image containing Albedo, normal, gloss, and spec. reflection. Further, we propose a color editing module that allows users to easily change hues of certain semantic regions of the final texture such as skin, lips, eyebrows, and tongue colors.}
    \label{fig:method}
\end{figure*}
We test few of the above approaches like modelling only PCA based meshes by calculating percentage of uniqueness within the generated samples along with novelty in respect to training data. We evaluate GAN based geometry generation method using shape maps as input and VAE based methods to generate geometry by directly regressing vertices. We observed following limitations.
\paragraph{Linear Shape Generation Methods} The geometry generated from popular morphable models  \cite{gerig2018morphable, li2017learning} doesn't guarantee diversity and novelty. Table \ref{tab:compare} compares the contemporary methods like DECA \cite{feng2021learning}, FLAME, and BFM with respect to diversity and novelty metrics along with other metrics that quantify the performance of these methods. Such methods also do not encourage entanglement or disentanglement features i.e, little to no control over semantics, thereby limiting their applications in artistic use cases like game character development and avatar creations. 
\paragraph{Non-linear Shape Generation Methods} It is common knowledge GANs are not feasible methods that a game artist can control whereas, Auto Encoders are good for reconstruction but lack semantic control over intermediate latent representation.  Therefore, both solely GAN based and AE based architectures are unsuitable for our objectives. Taher~\etal~ \cite{taherkhani2022controllable} use Auto Encoder to improve over reconstruction loss and sample real data  followed by two separate GANs to model geometry and texture in an disentangled representation. We use their code to train over our real data and generated new identities. From Table \ref{tab:compare}, we could quantitatively analyse the poor performance of their model in generating diverse and novel meshes. 
 Although image based GANs research approach taken by \cite{moschoglou20203dfacegan, gecer2020synthesizing, murphy2021artist, wood2021fake, li2020learning} do not align with our objective, we evaluated the quality and diversity of meshes generated from \cite{gecer2020synthesizing, li2020learning}. The results are discussed in Sec. \ref{sec:train} and Tab. \ref{tab:compare}. Intuitively the observed results show the loss in performance of the methods due to the process through which these shape maps are mapped from 3D to 2D representation. They don’t accurately represent the geometric details of the mesh due to the interpolations done in the process of creating shape maps. 
\paragraph{Texture} Most of prior efforts focus on direct synthesis of textures without humans-in-the-loop. However, in reality, game artists should still have control over the color of skin, and other facial features appearing in texture. Few approaches like \cite{murphy2021artist, murphy2020appearance, li2020learning} tried to introduce entanglement between texture and shape by jointly learning using linear methods and also offer control via user inputs. Li~\etal~\cite{li2020learning} on the other hand does not offer any explicit control over physical and demographic attributes of the generated identities and textures.
\section{Methodology}
\label{sec:meth}
We aim to build a method that can automatically synthesize high-quality 3D heads with a large of user controllability at every step of the synthesis process. 
In summary, our proposed method consists of three sequential modules (see Fig.~\ref{fig:method}): (1) Geometry Generator ($\Geom$), (2), Render Maps Generator ($\Render$), and (3) Color Transformer ($\CT$); each model is conditioned on outputs of the previous one.
\subsection{Shape Generator}
We define the mesh geometry\footnote{We use the terms "shape" and "geometry", interchangeably.} of our dataset as $S = {V, F}$ where $V\in \Re^{n\times3}$ is a set of $\bn$ vertices in $\bx, \by, \bz$ plane, and $F \in \Re^{\Gamma\times 3}$ are its faces represented by triangular polygon. We have registered our meshes to a common topology so $U$ and $F$ are consistent across entire dataset and only the vertices $V$ vary in the 3D space giving the mesh its shape and identity. These vertices have point-wise correspondence with other meshes in the dataset.

Following dataset registration, we annotate each mesh according to categories such as race, age, and gender. We then calculate mean mesh $\bar S_{mean}$ from the training set.  Following that we apply PCA over the complete training set and calculate the components (Eigen vectors) that bring the most variations sorted by their Eigen values. The offset values are calculated for game artist given controls like age, gender and race. For example, say the artist gives the values for race as ``asian", gender as ``male", and age as ``old". Then, mathematically we calculate, 

\begin{equation}
    \Delta c_{(g,a,r)} = \frac{1}{\mid S_{(g,a,r)}\mid} * \sum_i(S_{(g,a,r)}^i-\bar S_{mean})
\end{equation}
Here, $\Delta c_{(g,a,r)}$ is the offset that make the instance correspond to specific input controls $g$ gender, $a$ age, and $r$ race provided by the artist/user,  $S_{(g,a,r)}^i$ is the $i^{th}$ geometry or mesh in the training set that fall into the category of $g$. $a$, and $r$, and $\mid S_{(g,a,r)}\mid$ represents the cardinality or the number of meshes in the set $S_{(g,a,r)}$. 

After offset calculation, the new meshes are generated from PCA using the first $\mid \Vec{\beta} \mid$ principal components. The amount of variance represented by each principal component $\Vec{\beta} \in \Re^{3n \times \mid \Vec{\beta} \mid }$ is given by the coefficients $\Vec{\alpha} \in \Re^{1 \times \mid \Vec{\beta} \mid}$. To generate new meshes or geometry,  coefficient $\Vec{\alpha}$ is multiplied by weights obtained from random normal distribution in the same dimension to generate new coefficients. Mathematically, the linear model for geometry can be defined as, 

\begin{equation}
    w_i \sim \mathcal{N}(\mu,\,\sigma^{2}, i) \times \Vec{\alpha_i}    
\end{equation}
\begin{equation}    
    Geom_{new} = \bar S_{mean} + \sum^{\mid \Vec{\beta} \mid}_{i=1} w_i \beta_i
\end{equation}

where $w_i$ is the weight coefficient bringing variation in the principal directions defining the training set, $\mu$ and $\sigma$ are 0 and 1 for the normal distribution $\mathcal{N}$, for drawing $i^{th}$ principal component.

The generated mesh sample from PCA is then linearly displaced by $\Delta c_{(g,a,r)}$ to produce game artist desired mesh with specific age, race, and gender.
\begin{equation}
    \hat Geom_{(g,a,r)} = Geom_{new} + \Delta c_{(g,a,r)}
\end{equation}

From PCA, we get the diversified set of meshes represented by $\hat Geom_{(g,a,r)}$, with high fidelity. 
The generated set, is then converted into shape maps also called as position maps. Shape maps are the representation of 3D geometry in 2D plane. They are formed by interpolating the 3D vertices $V$ as r, g, b values and plotting their values in UV plane at the coordinates described by the texture coordinates $T$ in the geometry. The complete process is described in \cite{gecer2020synthesizing}. 

\subsection{Render Maps Generator}
\label{subsec:tex}

\begin{figure*}[!t]
    \centering
    \subfloat[TBGAN~\cite{gecer2020synthesizing}]{\includegraphics[height=2.3cm]{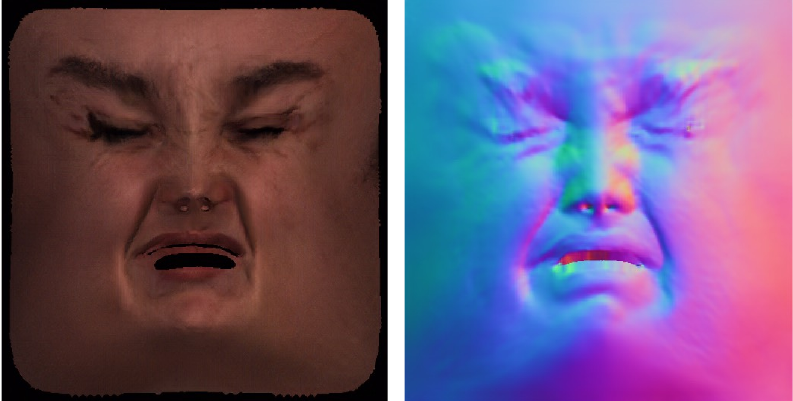}}\hfill
    \subfloat[MetaHuman~\cite{metahumans}]{\includegraphics[height=2.3cm]{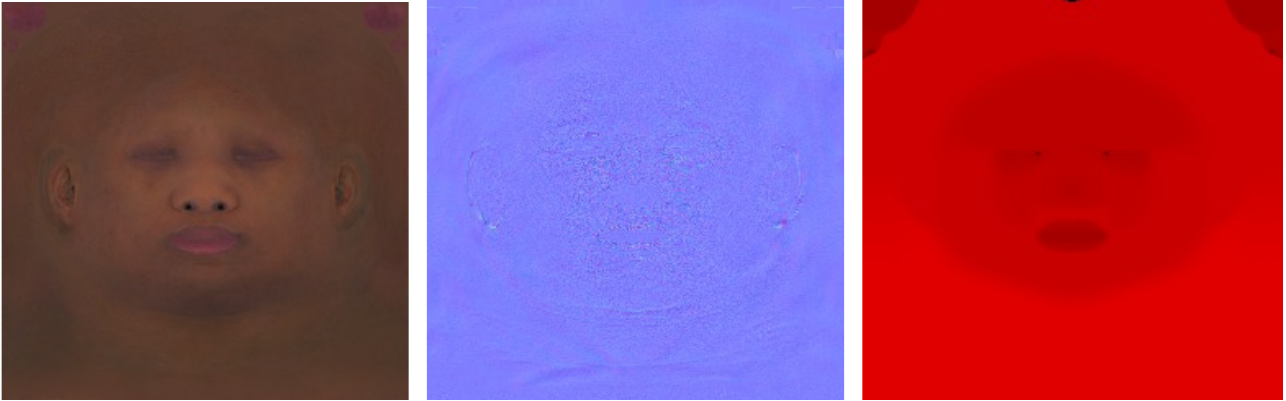}}\hfill
    \subfloat[DECA~\cite{feng2021learning}]{\includegraphics[height=2.3cm]{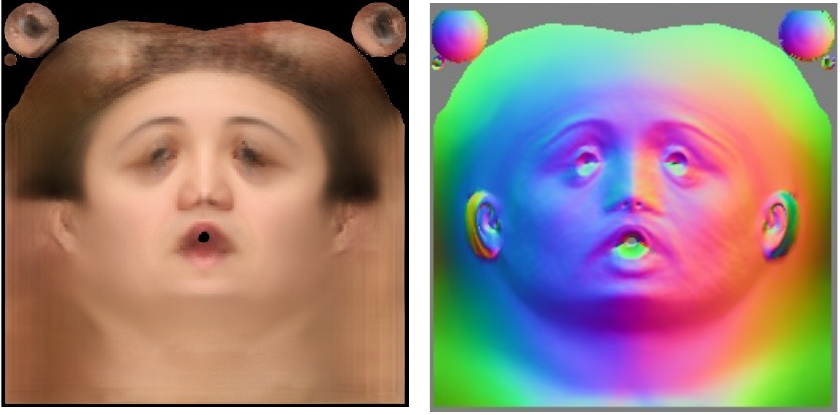}}\\
    \subfloat[Two examples of our render maps generated via $\Render$. In order: Albedo, Normal, Gloss, and Specular Maps.]{
        \includegraphics[width=0.5\linewidth]{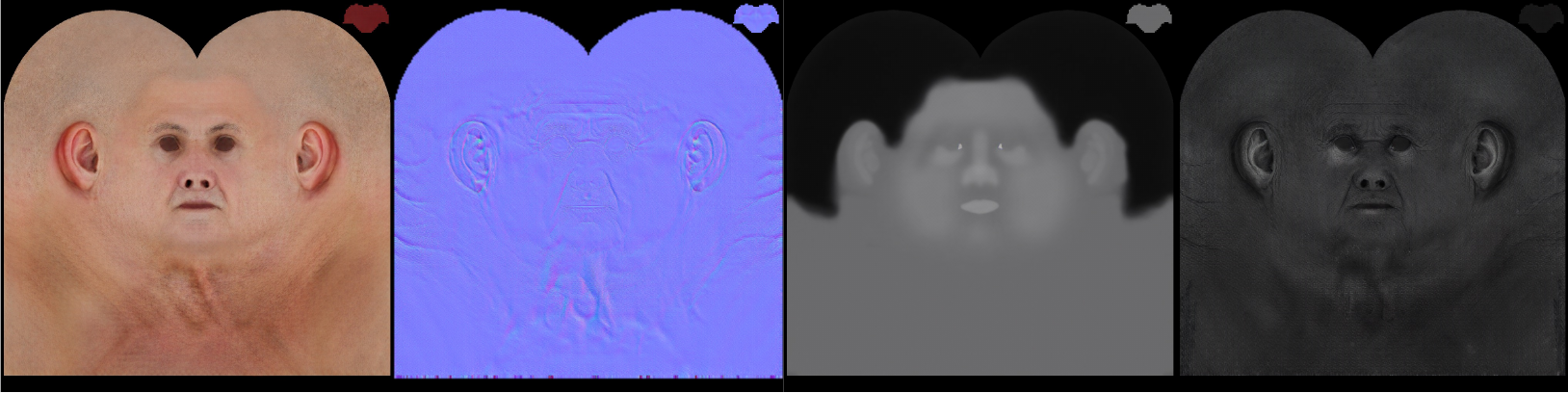}\hfill
        \includegraphics[width=0.5\linewidth]{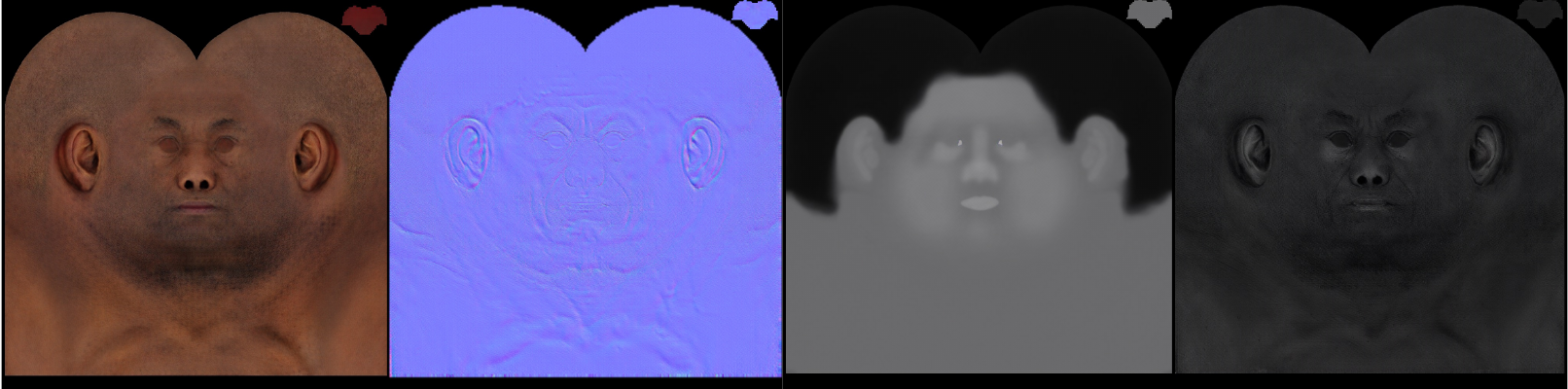}
    }
    \caption{Comparing render map synthesis of our method with respect to prevailing state-of-the-art methods. We output a larger number of render maps which directly improves perceptual quality.}
    \label{fig:textures}
\end{figure*}

The proposed render map generator, denoted as $\Render$, takes an input shape map image, $\bs \in \Re^{H\times W\times C}$, and outputs an $(N_{d}\times C_{d})$-channel image, $\Render(\bs)$, where $N_d$ are the number is the number of required render maps and $C_d$ is their respective channel-dimensionality. As the generator is conditioned only on the input shape map $\bs$, for a given head shape, the out render maps should be highly correlated with the head geometry. In this paper, we utilize $4$ render maps, namely, albedo, normal, gloss, and specular reflection with $3$ channels each. Therefore, the output of $\Render$ is a $12$-channel image. Figure \ref{fig:textures} (d) shows few examples from our $\Render$.
\paragraph{Visual Realism} A major requirement of any 3D head synthesis method that attempts to aid game artists in character design is high perceptual realism of the generated 3D heads. To achieve this, we employ multiple discriminators that have identical network structure operating at different scales, namely multi-scale discriminators. 
Specifically, we employ $3$ discriminators ($\ED^{\Render}_1,\ED^{\Render}_2, \ED^{\Render}_3$) and downsample the real and synthesized high-resolution images by a factor of $2$ and $4$ and train the generator via commonly employed adverarial GAN loss ($\EL^{\Render}_{GAN}$) \cite{isola2017image}.
To further improve realism, we incorporate a feature matching loss ($\EL^{\Render}_{FM}$) on the discriminator and a perceptual loss ($\EL^{\Render}_{P}$) via a pre-trained Convolutional Neural Network (CNN) as is common in image-to-image translation works \cite{isola2017image}.
\paragraph{Plausibility} The above losses encourages the synthesized render maps to come close to the real distribution. However, we also would like to ensure that the synthesized render maps follow the demographic attributes ($\ba$) desired by the user. We one-hot encode the attributes for the $3$ demographic groups $\{Gender, Age, Race\}$ such that each attribute is denoted as $1/0$ for with/without it. Our ground truth attributes include $2$ genders, $3$ age groups, and $4$ races. For \eg, one-hot encoding of a female that is young and caucasian will be $ba = [0, 1, 1, 0, 0, 0, 1, 0, 0]$ for $9$ possible attributes in the order of \{male, female, young, middle, old, asian, caucasian, african-american, mixed\}.
We find that the input shape maps contain enough demographic information such that $\Render$ usually outputs appropriate render maps belong to the user-chosen cohort and do not require explicit user-inputs. However, in order to further constrain this criteria, we propose an attribute classifier, $\EC$ which encourages the synthesized render maps to correctly own the desired demographic attributes, \ie $\EC(\br) \rightarrow \ba$. The attribute classifier is a CNN that is trained via classification loss:
\begin{equation*}
\EL^{\EC}_{cls} = E_{(\br, \ba)}[-\log(\EC(\ba | \br))]
\end{equation*}
while the generator attempts to output render maps belonging to the correct demographic group via,
\begin{equation*}
\EL^{\Render}_{cls} = E_{(\bs, \ba)}[-\log(\EC(\ba | \Render(\bs)))]
\end{equation*}
\subsection{Color Transformer}
We know that within a demographic population, there can be an indefinite number of skin color variations. To address this, our method allows for a second stage of editing towards the desired skin color texture within a specific demography. On the contrary, while previous methods allow for the generation of demography-specific textures \cite{murphy2020appearance}, they still do not allow for further editing of the Albedo map in a manner where users can easily obtain textures with semantic coloring of their choice. Consequently, artists spend significant efforts manipulating the Albedo map.

In other words, if a game artist prefers a certain texture synthesized by a GAN, prior studies are not robust enough to allow control over skin tone editing after the first stage of synthesis. To explain the importance of this module in our work clearly, we argue that while generative mesh models are helpful in quickly generating demographic-specific meshes, it is also imperative that control over texture parameters that match the demographic characteristics, along with the artist's choice, is included in the total solution.

To improve upon this lack of basic control over texture color, our proposed method utilizes a Color Transformer module that allows for changing the skin-tone, color of the eye-brows, lip and tongue color to any shade as picked by the game artist and therefore, provides for an unprecedented amount of diversity and flexibility in the final synthesized Albedo maps. Similar to the proposed Render Maps Generator ($\Render$), we model the problem of editing colors in a synthesized Albedo map as an image-to-image translation task. That is, we would like to obtain a function, say $\CT$, that takes a synthesized texture ($\Render(\bs)$) and a user-desired color palette ($\CM$) as inputs, and outputs an edited Albedo map $\CT(\Render(\bs)_{albedo}, \CM)$ that has: (a) the same identity content as $\Render(\bs)$, and (b) semantic coloring (such as lip, eyebrows, tongue, and skin colors) from $\CM$.
\paragraph{Controllability} We first need an easy way to encode the user-desired semantic color information which can then be translated to the Color Transformer module. A straightforward method would be to either trivially provide one-hot encoding attributes such as ``dark hair" or  ``brown skin" as input to the network, or take an RGB vector denoting exact color values desired by the end-user. However, this makes learning $\CT$ difficult as the network may not have any spatial cues as to which pixels in the Albedo maps requires editing. To this end, we propose utilizing a Semantic Coloring Map ($\CM$) where users can simply change the desired color in semantic regions of the face. 

We note that meshes that fall under a common topology follows the same UV space. We exploit this fact to first, manually build a semantic segmentation mask $\bS$ where we trace over Albedo maps output by $\Render$ and assign class labels to pixel regions falling under semantic regions. Specifically, we mark the lips, eyebrows, skin, and tongue. Then we use the mask to compute the median colors in all training examples, to get final Semantic Coloring Mask for each training instance\footnote{There are numerous methods for computing the dominant color value in image regions, however, we find that median works well in practice and has also been demonstrated to be effective with neural networks~\cite{kips2020gan}}. We just used single color for the entire face (target median colour) since albedo maps just represent the base color or diffuse color of the subject. The factors that define the realism in the texture like specular highlights, shadows, or surface details are controlled by other Render maps in our method (normal, gloss, and specular reflection maps). 
\paragraph{Color Transfer} Now, we need a way to enforce the $\CT$ network to learn the semantic color transfer from a source albedo map to a desired color map. If we only input the source Albedo map with its own corresponding color map, $\CT$ will fail to inherit any color transferability as there is no new information. Hence, we propose a random target shuffling strategy for training $\CT$. Given a mini-batch containing a set of corresponding albedo and Semantic Coloring Mask pairs,
$\{{\mathcal{G}_R(\mathbf{s})}_{albedo_i}, {\CM}_i\}$, we create random training tuplets: $\{({\Render(\bs)}_{albedo_i} , {\CM}_j, {\Render(\bs)}_{albedo_j})\}$, where $j$ may or may not be equal to $i$. We denote ${\Render(\bs)}_{albedo_i}$, ${\CM}_j$, ${\Render(\bs)}_{albedo_j}$ as the source albedo map, target color map, and target albedo map, respectively. 

 
We use reconstruction loss to encourage $\CT$ to transfer colors from source to target. 
\begin{equation*}
\EL^{\CT}_{rec} = || \CT\left({\Render(\bs)}_{albedo_i}, {\CM}_j\right) - {\Render(\bs)}_{albedo_j} ||_2
\end{equation*}
Note that $\CT$ does not have access to any identity-related features of the target albedo since the input to the network is the source albedo map and semantic colors of the target albedo. Due to this, the network only learns to transfer the color information present in the source albedo rather than any content-related features. Also, we find that a weaker constraint of allowing $j = i$ (source and target albedos are the same) leads to better convergence as long as the probability of this happening is kept low by introducing a larger batch size.
\paragraph{Visual Quality} To maintain the visual quality of the synthesized results after semantic color transfer, we introduce the same losses as previous step ($\Render$). That is, encouraging visual realism to synthesized outputs by employing (i) adversarial loss via multi-scale discriminators $\EL^{\CT}_{GAN}$, (ii) feature-matching loss $\EL^{\CT}_{FM}$, and (iii) perceptual loss $\EL^{\CT}_{P}$.
\subsection{Training Framework}
We train our proposed user-controllable, diverse, and high-quality 3D head synthesis method in a sequential manner in 2-stages (see Fig.~\ref{fig:method}). We first train our Render Maps Generator $\Render$ with the following objectives:
\begin{align*}
    \begin{split}
    &\min_{\Render}\EL_{\Render} = \EL^{\Render}_{GAN} + \lambda_{f}(\cdot\EL^{\Render}_{P} + \cdot\EL^{\Render}_{FM}) + \lambda_{cls}\cdot\EL^{\Render}_{cls},
    \end{split}\\
    &\min_{\ED^{\Render}}\EL_{\ED^{\Render}} = -\EL^{\Render}_{GAN},\\
    \begin{split}
    &\min_{\EC}\EL_{\EC} = \EL^{\EC}_{cls}.
    \end{split}
\end{align*}
After $\Render$ is trained to output visually realistic and plausible render maps from input shape maps, we then train $\CT$ with the following aim:
\begin{align*}
    \begin{split}
    &\min_{\CT}\EL_{\CT} = \EL^{\CT}_{GAN} + \lambda_{rec}\cdot\EL_{rec} + \lambda_{f}(\cdot\EL^{\CT}_{P} + \EL^{\CT}_{FM}),
    \end{split}\\
    &\min_{\ED^{\CT}}\EL_{\ED^{\CT}} = -\EL^{\CT}_{GAN}.
\end{align*}
\section{Experimental Results}
\label{sec:train}
We acquire a dataset consisting of head scans of $104$ diverse subjects from 3DScanStore \cite{3dscanstore}. See Supp. for dataset statistics. All meshes are re-topologized in a common topology. We divided our real data into training and testing set in 9:1 ratio. After fitting the $94$ meshes (further referred as real/train data) in the proposed PCA space, we synthesize $30,000$ new meshes for further experiments.
We use ADAM optimizers in PyTorch with for both render maps and color transfer networks. Empirically, we set $\lambda_f = 10.0, \lambda_{cls} = \lambda_{rec}  = 1.0$ More details in Supp.
\begin{table}
\begin{minipage}{\linewidth}  
    \centering
    \caption{Comparison over quantitative metrics between state-of-the-art linear, non-linear, and human curated mesh generation frameworks. Here D stands for Diversity, U for Uniqueness, S for Specificity, N for Novelty, and P for Performance }
    \label{tab:compare}
    \setlength\tabcolsep{-5pt}
    \begin{tabularx}{\linewidth}{@{\extracolsep{\fill}} l *{6}{c}}
    \toprule
         {\textbf{Method}}&{\textbf{D$\uparrow$}}&{\textbf{U \%}}&{\textbf{S$\downarrow$}}&{\textbf{N \%}}&{\textbf{P\%}}  \\
         \midrule  
         \midrule
         \multicolumn{6}{c}{\large Non-linear generative methods} \\
         \midrule
         
         TBGAN \cite{gecer2020synthesizing}& 145.12&22.7&146.66&14.9&7.8\\         
         ICT\cite{li2020learning}&28.75&25.3&28.89&19.5&12.8\\        
         DECA\cite{feng2021learning}&43.47&15.8&45.33&11.8&7.0\\        
         DAD-3DHeads\cite{martyniuk2022dad}&1.52&27.5&1.5&22.3&17.4\\
         3DFaceCam~\cite{taherkhani2022controllable}&41.19&27.3&43.5&20.6&15.2\\
    
         \midrule
         \multicolumn{6}{c}{\large Human curated meshes} \\
         \midrule
     
         Metahuman \cite{metahumans} &19.65&-&-&-&- \\
        
         3DScanStore (ours) \cite{3dscanstore}&30.89&-&-&-&- \\
       
         \midrule
         \multicolumn{6}{c}{\large Generative linear models (PCA/LBS)} \\
         \midrule
         BFM \cite{gerig2018morphable}&86.09&30.5&85.75&24.7&17.3\\
        
         FLAME \cite{li2017learning}&9.85&25.8&9.82&18.8&12.3\\
         MetaHuman \cite{metahumans}&20.11&24.8&19.69&18.6&11.3\\
        \midrule
         Ours \footnote{\label{note1} PCA generated test data} & 68.76&26.3&69.13&17.61&10.6\\         
         Ours (with real test data)& 62.3&74.4&46.13&74.4&66\\
    \bottomrule    
\end{tabularx}
\end{minipage}

\end{table}
\subsection{Analysis of our Shape Generator: Diversity in Shapes}
Our first goal of the 3D head synthesis method is to be able to create diverse and unique heads. To this end, we evaluate our shape generation method rigorously under both qualitative and quantitative settings. All meshes for baselines and ours are normalized between $[-1, 1]$.

To compare our Shape Generator with recent methods, we use $10$ meshes in test set. To create the generated set, we synthesize $1000$ samples. For quantitative evaluation of our PCA generated mesh we compare different state-of-the-art methods in shape generative modeling encompassing linear, non-linear, and also human curated domain. For the same, we came up with five metrics like Diversity, Specificity, Uniqueness, Novelty, and Performance. The brief description and formulation of each metric is explained below. 

We have used Euclidean distance to calculate inter and intra distance between generated and real test samples. To calculate the mean Euclidean distance ($E$) between N samples in a 3D space and M samples in another 3D space, we use the following formula: 

\begin{equation}
E = \sqrt{\sum_{i=1}^{V} \sum_{j=1}^{3} (A_{ki,j} - B_{li,j})^2}
\end{equation}

Here, 
\begin{itemize}
    \item \textit{Euclidean Distance} represents the Euclidean distance between a mesh from matrix A of generated samples and a mesh from matrix B of real test samples.
    \item \textit{k} and \textit{l} are indices representing the respective samples from matrix A and matrix B, ranging from $1$ to $nSamples$ for generated and $1$ to $nReal$ for real test samples.
    \item \textit{i}  ranges from 1 to \textit{V}, representing the number of vertices in a sample mesh.
    \item \textit{j} ranges from 1 to 3, representing the dimensions of the vertices.   
    \item $A_{ki,j}$  represents the element of row \textit{i}, column \textit{j} of matrix A for the k\text{th sample}
    \item $B_{li,j}$  represents the element of row \textit{i}, column \textit{j} of matrix B for the l\text{th sample}.
    
\end{itemize}
When there is intra distance calculation, we use the notation of euclidean distance $E_{intra}$ where matrix A and B of samples are equivalent (either representing generated samples $E_{intra_{gen}}$ or real test samples $E_{intra_{real}}$). For inter distance between real and generated samples the notation is $E_{inter}$.

\subsubsection{Diversity} quantifies the difference in generated meshes (samples) that is how diverse generated meshes are from each other so that they represent the complete data domain and the model does not collapse. We have formulated Diversity as : 

\begin{equation}
    Diversity = \frac{1}{n} \sum_{i=1}^{n} \left(\frac{1}{m} \sum_{j=1}^{m} E_{intra_{gen}}[i][j]\right)
\end{equation}
where, \textit{n} and \textit{m} are the number of samples (nSamples) where n is equal to m, $E_{intra}$represents the pairwise Euclidean distance between the i-th and j-th samples ($i < j$) in the generated meshes.

For a single method, we compute pairwise euclidean distance between all samples. This generates a matrix of $(nSamples\times nSamples)$. We then calculate the mean across columns $j$ to get the vector of size $nSamples$. Finally, we compute the total mean across $nSamples$ again to get a diversity value.  The smaller the diversity value, the more similar the generated samples are to each other, indicating lower diversity. Conversely, a higher diversity value suggests that the generated samples are more dissimilar, which is often desired for diverse and representative sample sets. Thus, Diversity checks that "mode collapse" does not happen.

\subsubsection{Specificity} measures the closeness of the generated and real distribution also commonly known as "fidelity" of the model. This quantifies the quality of the generated meshes. Since even synthesized meshes that are out-of-distribution will bring about large diversity within the sample set, we posit that it is important to consider both diversity \emph{and} specificity. Specificity is calculated by taking the average inter-distance between each pair of real and generated sample. Here, lower specificity indicates higher quality. The given formula computes the specificity metric by averaging the inter-distances between the generated meshes and the real test samples. 
\begin{align}
\text{Specificity} &= \frac{1}{nSamples} \sum_{i=1}^{nSamples} \left(\frac{1}{nReal} \sum_{j=1}^{nReal} E_{intra_{real}}[i][j]\right)
\end{align}
where, \textit{nSamples} and \textit{nReal} are the number of data-points in generated and real test dataset, respectively. $E_{inter}$ represents the pairwise Euclidean distance between the i-th and j-th samples ($i < j$).

\subsubsection{Threshold $\tau$}: The value of threshold $\tau$ will be different for each method. We define threshold as the mean of the minimum matching distance between all pairs of real dataset. The real dataset is assumed to be have maximum diversity therefore, the minimum matching distance between the real samples should be a value in the real distribution from which which all inter and intra distances should be higher to ensure uniqueness and novelty. The steps to calculate $tau$ is:
\begin{itemize}
    \item Calculate intra distance between real test samples. $E_{intra_{x}}$ where \textit{x} represent the real test samples.
    \item Set the elements in lower triangle of $E_{intra_{x}}$ to infinity so that the distances between same samples are not repeated.\\
    $\forall i \leq j, \quad E_{intra_{ij}} = \infty$
    \item Calculate the threshold as the mean of the minimum values for each row of $E_{intra_{ij}}$. 
    \begin{equation}
             \tau = \frac{1}{nReal} \sum_{i=1}^{nReal} \min_{j \neq i} \left(E_{intra_{real}{ij}}\right)
    \end{equation}
\end{itemize}    
    where, $nReal$ is the number of real test samples $x$, $i$ and $j$ represent sample indices, and $min_{j \neq i}$  calculates the minimum value in each row while excluding the diagonal values (i.e., comparing each sample to others excluding itself).

\subsubsection{Uniqueness}: measures the ratio of generated meshes that are different from other synthesized samples. It is computed by finding out how many pairs of meshes within the  generated set have their distances higher than a particular matching threshold $\tau$ over all the generated samples. The higher the uniqueness ratio, the more unique meshes we obtain from the particular approach. 
The formulation is given below:
\begin{itemize}

\item Calculate $\alpha$ based on minimum values in each row of $E_{intra_{gen}}$
\begin{equation}
   \alpha_i = \min_{j \neq i} \left({E_{intra_{gen}ij}}\right) \geq \tau
\end{equation}
\item Calculate \textit{Uniqueness} as the percentage of $\alpha$ values that meet the threshold $\tau$:
\begin{equation}
    Uniqueness = \frac{\sum_{i=1}^{nSamples} \alpha_i}{nSamples} \times 100
\end{equation}
\end{itemize}
\subsubsection{Novelty}: Specificity has its limitations in defining the novelty in generated meshes. For instance, trivially overfitting to meshes present in the training set will lead to very low specificity. Novelty justifies that the approach of mesh generation is providing new and useful 3D heads to users. Therefore, we define \textit{Novelty} as a measure that gives us a ratio of meshes that differ from the real test set. It is calculated by measuring how many pairs of generated and real samples have distance greater than a certain match threshold $\tau$. 
The formulation is given below:
\begin{itemize}

\item Calculate $\beta$ based on minimum values in each row of $E_{inter}$
\begin{equation}
   \beta_i = \min_{j} \left(E_{inter_{ij}}\right) \geq \tau
\end{equation}
\item Calculate \textit{Novelty} as the percentage of $\beta$ values that meet the threshold $\tau$:
\begin{equation}
    Novelty = \frac{\sum_{i=1}^{n} \beta_i}{n} \times 100
\end{equation}

\end{itemize}
where, $n$ is the number of pairs in generated and real set whose distance is greater than $\tau$
\subsubsection{Performance}: As we discussed, we need to consider both Uniqueness and Novelty when evaluating the utility of any proposed 3D head synthesis method. Therefore, in an effort to unify the two metrics, we define the overall performance of the taken approach by averaging uniqueness and novelty.
\begin{equation}
    to\_keep_i = \alpha_i \land \beta_i 
\end{equation}
\begin{equation}  
Performance = \frac{\sum_{i=1}^{m} to\_keep_i}{m} \times 100
\end{equation} 
where, $m$ here is the number of samples who are both unique and novel. 

While we used our real scans as a testing set to compute all metrics, obtaining real scanned datasets from comparative methods like \cite{beeler2010}, \cite{ranjan2018generating}, VOCA \cite{VOCA2019} and D3DFACS \cite{cosker2011facs} proved challenging\footnote{FaceScape and \cite{beeler2010} are not available, while CoMA \cite{ranjan2018generating}, VOCA \cite{VOCA2019} and D3DFACS \cite{cosker2011facs} are trained on only 10-12 3D scans.}. Therefore, we chose methods that are more recent and have similar objective. We used a percentage of synthesized meshes from these methods as a representation of their training data. For fair comparison, the same experiment is repeated for our method. Note that linear methods like PCA is an interpolation of the source datasets. Therefore, if the training data does not have enough diversity, the same could not be reflected in the PCA generated meshes. 
In case of Non-linear methods, even if these models are trained on large training datasets, if not diverse and inclusive, will also be unable to generate diverse and novel meshes. Due to unavailability of real scans, majority of prior work fails to quantitatively validate their work by measuring such metrics. From the Table \ref{tab:compare}
 We note that although TBGAN has high diversity, it is also prone to high specificity. This can also be seen in its generated mesh quality (see Supp.). The magnitude of diversity in TBGAN can likely be attributed to incorporating expressions as compared to other methods which are evaluated on synthesized meshes with neutral expressions. A clear trend is observed: methods that model geometry in the linear space have higher performance in both uniqueness and novelty as compared to non-linear methods. We posit that this is likely a sign of overfitting due to limited available 3D head data.
From Tab. \ref{tab:compare}, we find that our approach to modeling shapes is far superior and also leads to visually appealing and plausible meshes. \emph{More examples for each of these methods are in Supp.}
\begin{figure}[!t]
    \centering
    \includegraphics[width=0.915\linewidth]{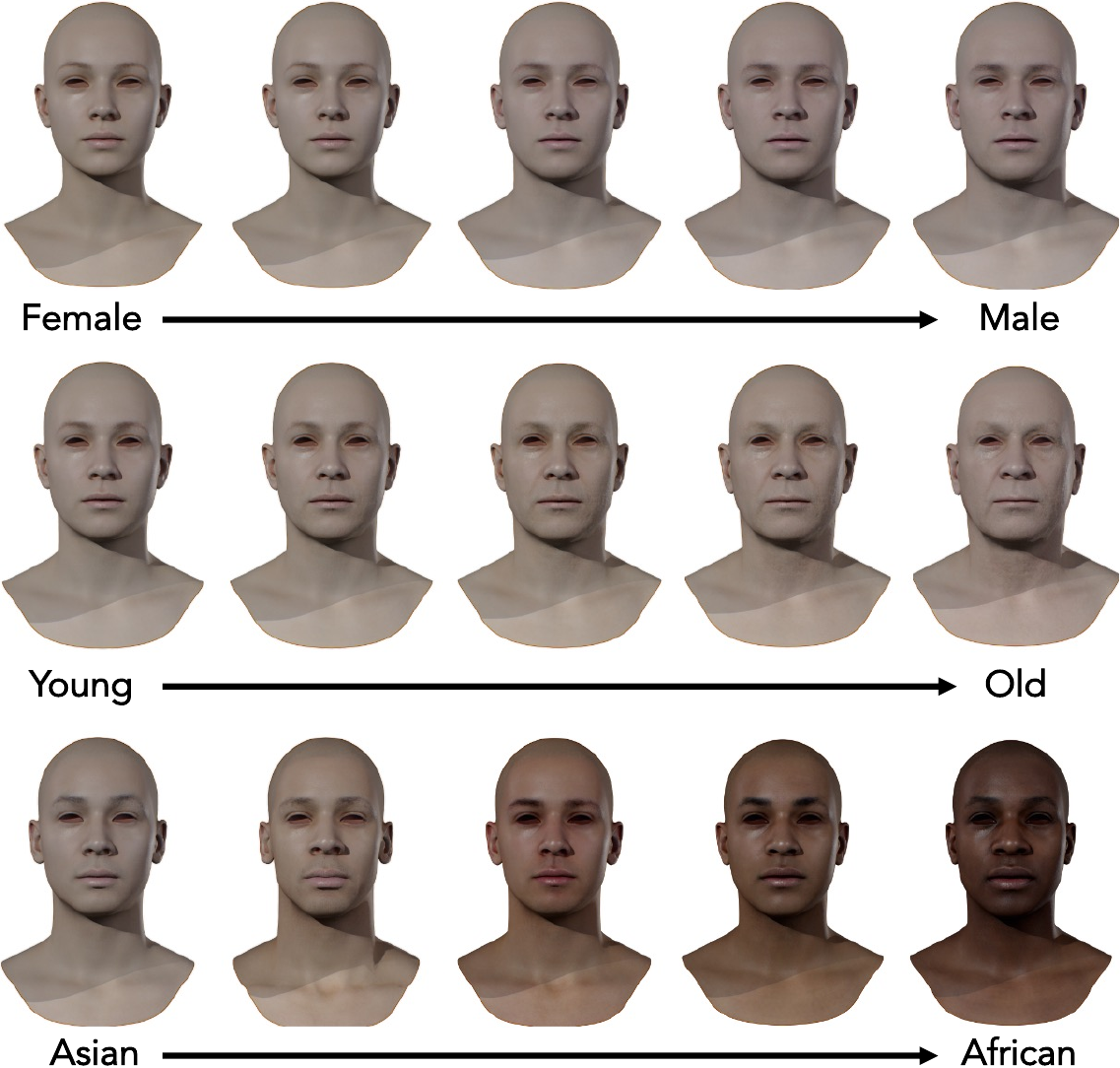}
    \caption{Each row demonstrates interpolation of a single attribute. Without explicitly inputting the user-defined choices, our proposed Render Maps Generator automatically infers correct user-inputs from shape maps alone. Also, the proposed Shape Generator maintains high disentangled between attributes.}
    \label{fig:interpolation_attributes}
\end{figure}
\subsection{Analysis between Shape and Render Maps}
As discussed earlier, geometry and render maps are deeply correlated in the physical nature and as such, methods attempting automated 3D synthesis should also follow suit. Our method vastly differs from majority of prior studies on this front since we directly condition synthesis of all render maps on geometric information present in shape maps. In this work, we show the entanglement between shape and render maps via analyzing the correlation between user-defined demographic attributes such as race, age, and gender. \emph{Note that all render maps are synthesized only from shape maps alone without any explicit demographic inputs to the Render Maps Generator $\Render$.}
\begin{table}[h]
    \centering
        \caption{Quantitative evaluation of entanglement between shape and texture shown through classifier accuracy per category (all in \%).}
    \label{tab:ent}
     \renewcommand{\arraystretch}{1.2}
    \centering
    \begin{tabular}{cccc}
         \textbf{Dataset}&\textbf{Gender}&\textbf{Age}&\textbf{Race}  \\
         \hline
         ShapeMaps from $\Geom$ & 96.15 & 87.5 & 91.35 \\
         
         \hline
    \end{tabular}

\end{table}
 
\begin{figure}[!t]
    \centering
    \includegraphics[width=\linewidth]{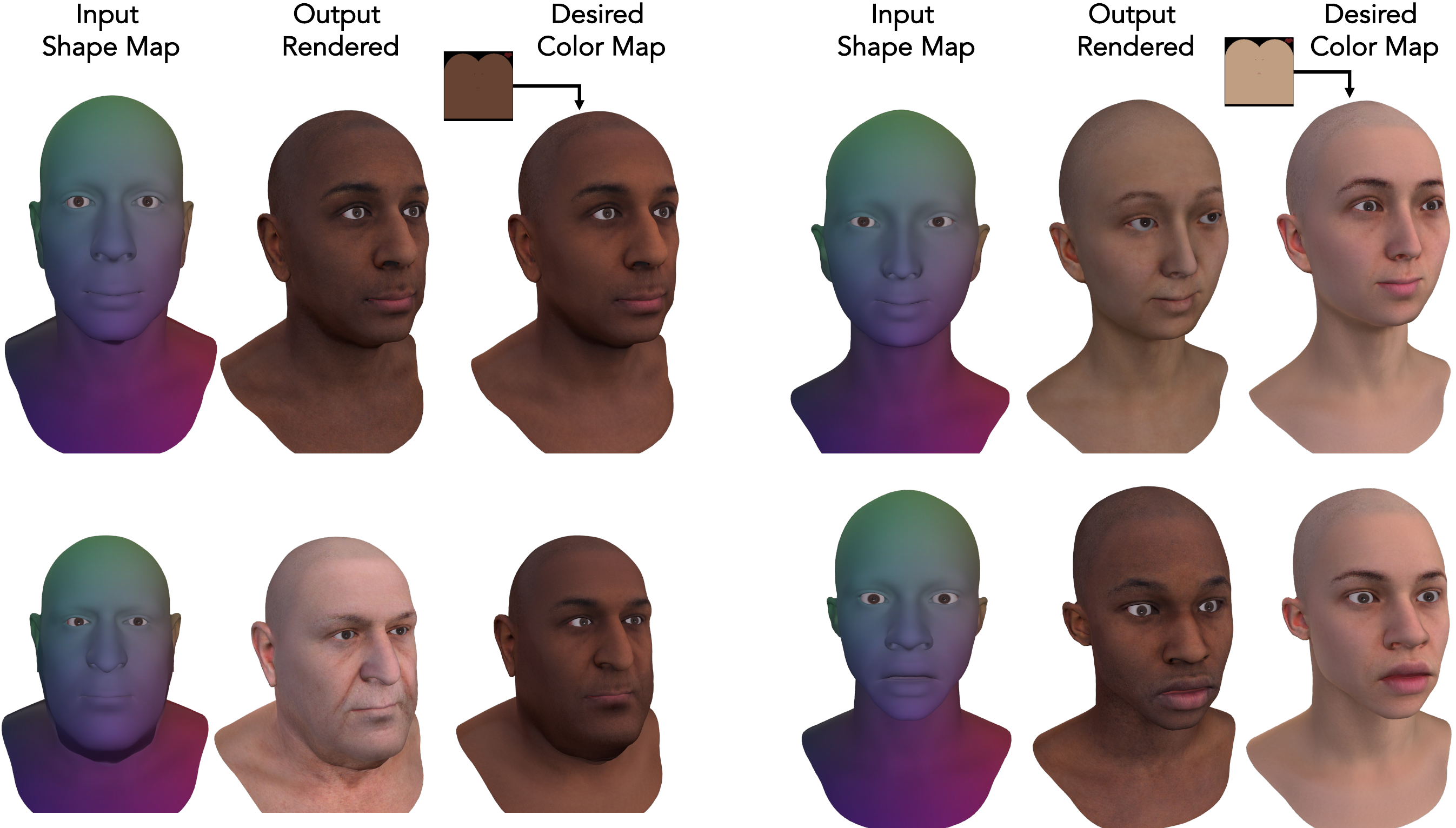}
    \caption{Four examples of plausible renders by our Render Maps Generator. In addition, we show the efficacy of the proposed Color Transfer model in editing semantic color changes such as skin, lips, eyebrows, and tongue colors, while maintaining the original identity of the render.}
    \label{fig:color_transfer}
\end{figure}
\begin{figure*}[!t]
    \centering
    \subfloat[AI~\cite{li2017learning} ]{\includegraphics[height=3cm]{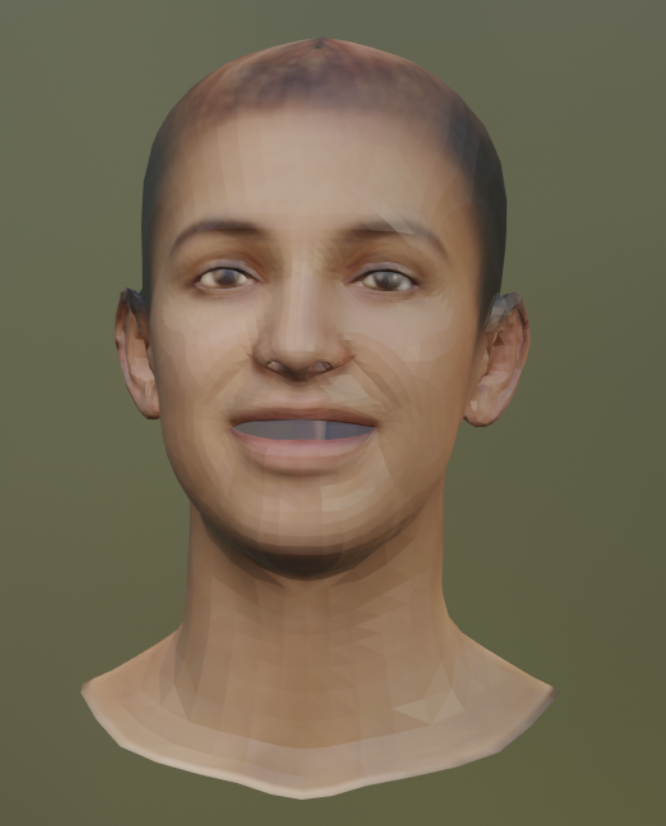}}\hfill
    \subfloat[AI~\cite{wood2021fake} ]{\includegraphics[height=3cm]{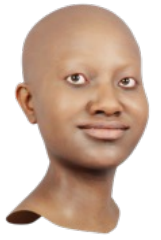}}\hfill
    \subfloat[AI~\cite{gerig2018morphable} ]{\includegraphics[height=3cm]{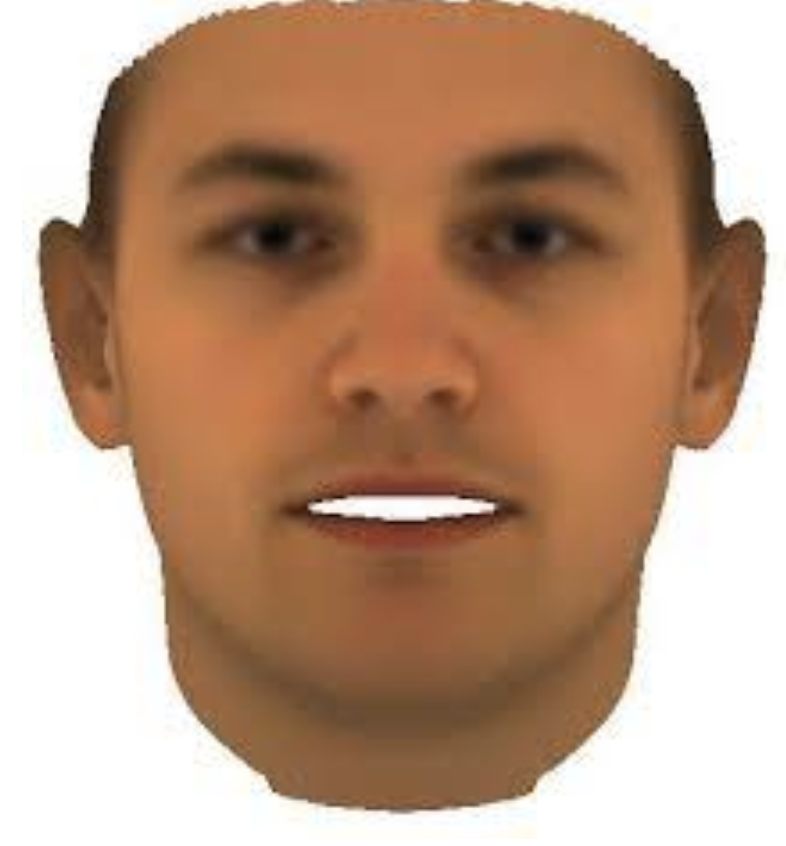}}\hfill
    \subfloat[AI~\cite{gecer2020synthesizing} ]{\includegraphics[height=3cm]{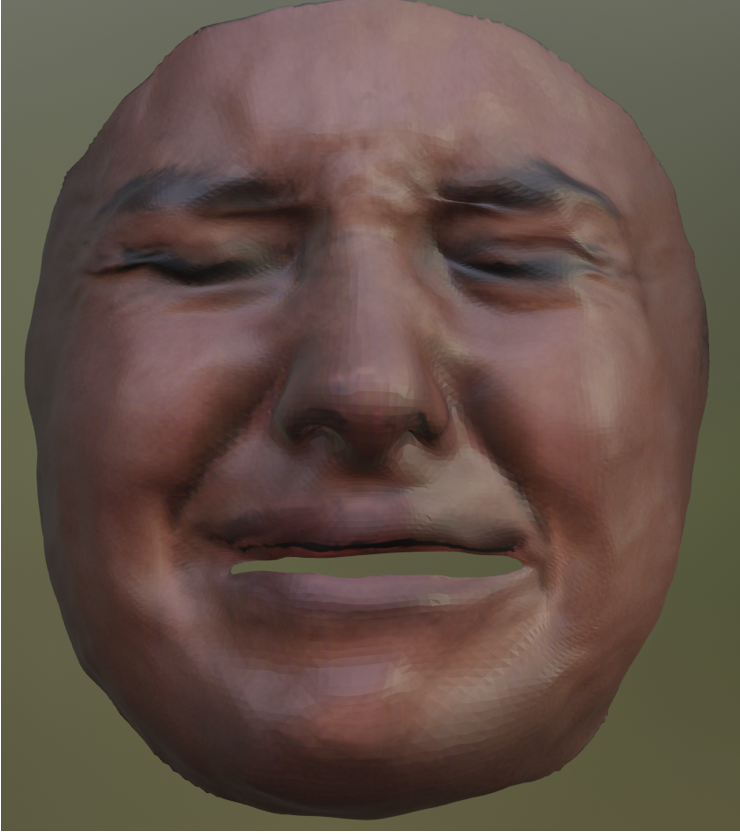}}\hfill
    \subfloat[Artist~\cite{synthesis_ai} ]{\includegraphics[height=3cm]{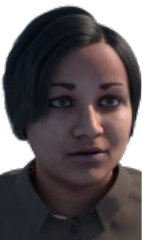}}\hfill
    \subfloat[Artist~\cite{metahumans}]{\includegraphics[height=3cm]{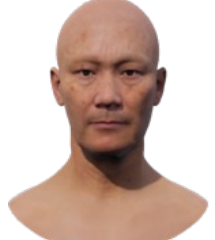}}\hfill
    \subfloat[Ours]{\includegraphics[height=3cm]{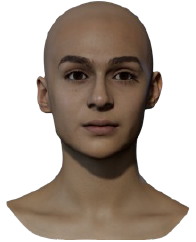}}\hfill
    \caption{Comparing our rendered 3D head with prevailing state-of-the-art synthesis methods. Most methods are either hand-crafted by game artist or sway too much towards synthetic domain. From (a) to (d) - AI synthesized meshes, (e) have heavy artist involvement, (f) is artist generated. In comparison, our proposed method maintains a high degree of visual realism. (\textit{comparing qualitatively from available render data only})}
    \label{fig:render}
\end{figure*}
We have compared the outputs of our $\Render$ with other state of the art methods that generate two or more maps in Figure \ref{fig:textures}. We also visualize the efficacy of the entanglement between shape and render maps of our proposed method in Fig. \ref{fig:interpolation_attributes}. We take the average PCA mesh $S_{mean}$ and linearly interpolate between two different PCA directions (say, $S_1$ and $S_2$) using a smoothing operator $\alpha$ via,  
 \begin{equation}
 \label{eq:ent}
    S_{out} = \bar S_{mean} + ((1 - \alpha) * S_{1} + \alpha * S_{2}) 
 \end{equation}
 \begin{figure}[!t]
    \centering
    \includegraphics[width=\linewidth]{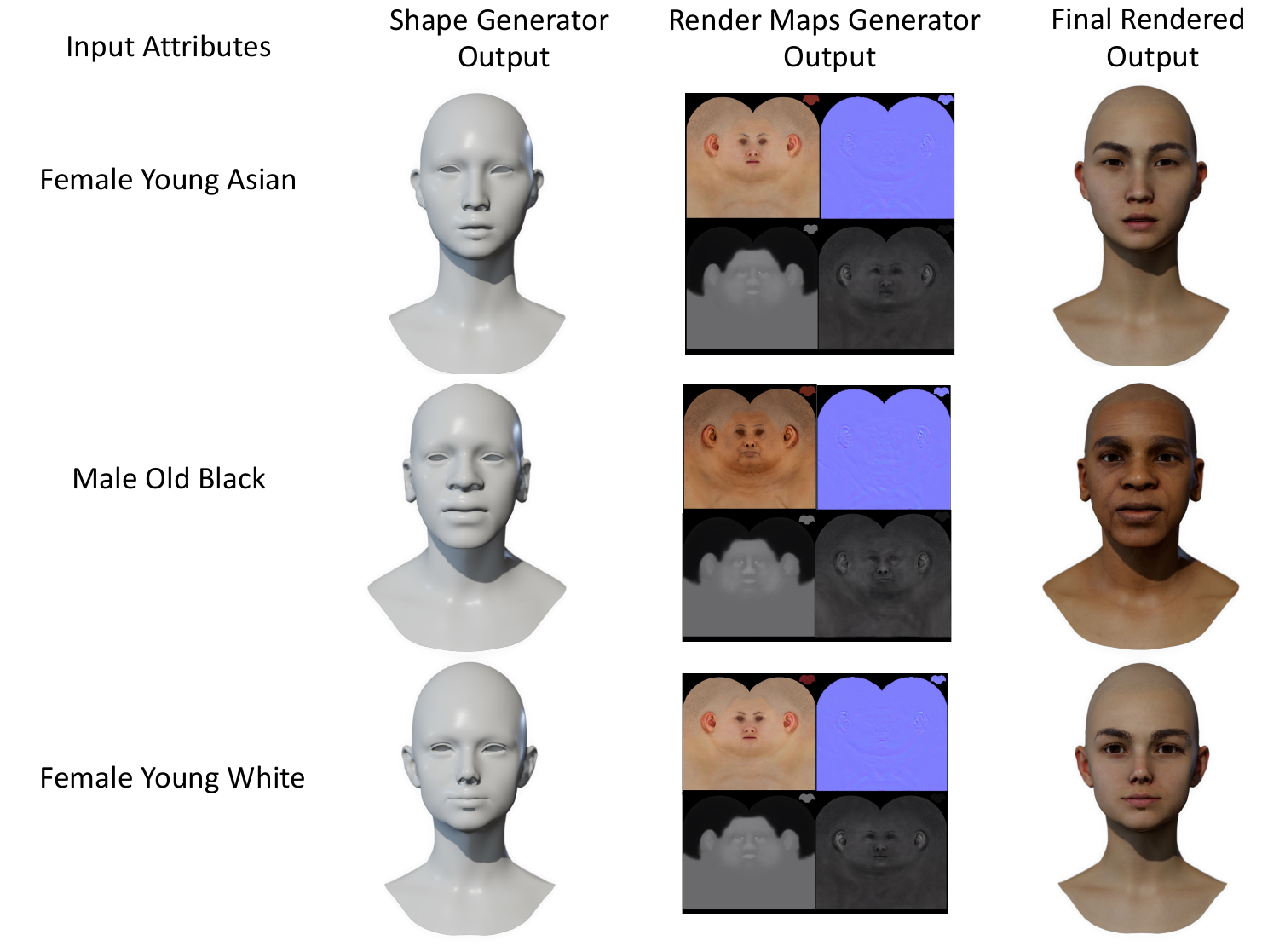}
    \caption{Qualitative examples representing outputs of $\Geom$ and $\Render$ given aspecific input demographic.}
    \label{fig:attr}
\end{figure}
For example, in order to study how our $\Render$ behaves to gender-specific user inputs, we can consider $S_{1}$ as the average principal components of females, and $S_{2}$ as average male components. Then, while we interpolate from females to males, we subsequently synthesize intermediate shape maps and obtain the resulting render maps from $\Render$. We see that our plausibility requirements are satisfied and that the render maps appropriately infers and owns the user-defined demographic directions from shape map alone. This is even more evident in the case of interpolating between different races where both shape and render maps react correctly to varying demographic directions.

We also report the quantitative performance of our correlation between synthesized head geometry and render maps. For this, we trained a classifier\footnote{Note that we also employ an attribute-classifier $\EC$ in the training process, however, the classifier used in our experiment here is a black-box classifier based on a different architecture (see Supp.)} based on a pre-trained ResNet18 architecture. The classifier is trained on real Albedo maps, each labeled into three categories, {$Race$, $Age$, and $Gender$}, where $Race = \{Asian, Black, Mixed, Caucasian\}$, $Age = \{Young, Middle, Old\}$, and $Gender = \{Female, Male\}$. We randomly samples $30,000$ meshes with random attribute directions in the PCA space. Via shape maps obtained from these meshes through $\Geom$, we then infer the corresponding render maps via $\Render$. We evaluate the attribute classification performance directly on these render maps, where ground truth labels are set when generating meshes via $\Geom$. From Table \ref{tab:ent}, we see that the attribute classifier can predict the ground truth demographics from the synthesized render maps via $\Render$ with high accuracy. We then conclude: (a) there is low domain gap between real and PCA meshes, and (b) our proposed $\Render$ achieves high correlation with user-defined demographic inputs. To the best of our knowledge, our method is the first to show such high level of correlation between shape and render maps both quantitatively and qualitatively. Fig. \ref{fig:attr} show the correlation between shape and render maps when specific attributes are given as input to the Shape $\Geom$ and Render Maps $\Render$ Generator.
\subsection{Disentanglement between geometry attributes}
Due to the linearity of our geometry modeling, we posit that we can achieve high levels of disentanglement between different user-defined demographic attributes. Due to this feature, it is easier to find controlled latent directions for each attribute and change one attribute while keeping the others the same. In contrast, modeling this in non-linear space is still an unsolved problem and requires large amounts of data; even then complete disentanglement is never guaranteed. In Fig. \ref{fig:interpolation_attributes}, we show the effects of this disentanglement. Each row demonstrates that while varying a single attribute, other demographic effects are not observed. For instance, varying gender shows little to no effect on race and age of the intermediate meshes. In this manner, our controllability requirements are also satisfied.
\subsection{Semantic Color Editing}
In Fig. \ref{fig:color_transfer}, we demonstrate our semantic color editing module. We first see that plausibility and perceptual realism in the synthesized renders by $\Render$. Next, we notice the perceptual accuracy in transferring color to the desired color map while changes in identity or visual quality is hardly noticeable. In addition, we also find that taking the desired color map as the median colors obtained from the input itself leads to nearly identical reconstruction. This again highlights the disentanglement identity-related content and semantic coloring. Fig. 6 in Supp. shows some more visual examples for better qualitative evaluation.
\subsection{Quality of our Rendered Heads} In Fig. \ref{fig:render}, we compare the rendered outputs of different baseline 3D head synthesis methods. We note that compared to all other methods which are (i) either hand-crafted by game artists that spent many months in crafting them, or (ii) synthesized towards synthetic data, our approach is able to maintain visual realism with unprecedented quality. See Supp. for more examples at much higher resolutions. Unfortunately, more recent methods like \cite{lattas2021avatarme++}, \cite{lattas2023fitme}, \cite{lin2022realistic}, \cite{lin2021meingame} do not provide open-source codes to output samples for qualitative comparison. Hence, evaluating more recent methods without inference codes and samples proved challenging. 
\section{Conclusion}
We proposed a new method of 3D head synthesis which take user inputs such race, age, and gender and automatically outputs diverse meshes with unprecedented quality. We show both quantitatively and qualitatively that our proposed method outperforms prevailing state-of-the-art in 3D head synthesis both in terms of diversity and perceptual realism. In addition, we show the our proposed Color Transformer can further allow users to change semantic color changes in final renders such editing skin, lips, eyebrows, and tongue colors. We plan to extend our work to model fine features such as scars, marks, and face tattoos.

\bibliographystyle{ACM-Reference-Format}
\bibliography{manuscript}
\end{document}